\documentclass[a4paper]{article}

\usepackage{INTERSPEECH2021}
\usepackage{times}
\usepackage{latexsym}
\usepackage{multirow}
\usepackage{amsmath}
\usepackage{amsfonts}
\usepackage{mathtools}
\newcommand{\tabincell}[2]{\begin{tabular}{@{}#1@{}}#2\end{tabular}}
\title{A Context-Aware Hierarchical BERT Fusion Network for Multi-turn Dialog Act Detection}
\name{Ting-Wei Wu$^1$, Ruolin Su$^1$, Biing-Hwang Juang$^1$}
\address{
  $^1$Georgia Institute of Technology}
\email{waynewu@gatech.edu, ruolinsu@gatech.edu, juang@ece.gatech.edu}

\begin{document}

\maketitle
\begin{abstract}
The success of interactive dialog systems is usually associated with the quality of the spoken language understanding (SLU) task, which mainly identifies the corresponding dialog acts and slot values in each turn. By treating utterances in isolation, most SLU systems often overlook the semantic context in which a dialog act is expected. The act dependency between turns is non-trivial and yet critical to the identification of the correct semantic representations. Previous works with limited context awareness have exposed the inadequacy of dealing with complexity in multiproned user intents, which are subject to spontaneous change during turn transitions. In this work, we propose to enhance SLU in multi-turn dialogs, employing a context-aware hierarchical BERT fusion Network (CaBERT-SLU) to not only discern context information within a dialog but also jointly identify multiple dialog acts and slots in each utterance. Experimental results show that our approach reaches new state-of-the-art (SOTA) performances in two complicated multi-turn dialogue datasets with considerable improvements compared with previous methods, which only consider single utterances for multiple intents and slot filling. 

\end{abstract}

%%%%%%%%%%%%%%%%%%%%%%%%%%%%%%%%%%%%%%%%%%%%%%%%%%%%%%%%%

\section{Introduction}

% Intro of task-dialog, tasks
With the recent success of service assistants such as Alexa, Cortana and Siri, research attempts to expanding spoken language understanding (SLU) applications are becoming ubiquitous~\cite{zhang2020recent}. In canonical task-oriented dialogs, SLU establishes so-called semantic frames by capturing semantics in terms of intents and slots from speech recognized utterances~\cite{weld2021survey}.  These intents specify goals which commit speakers to some course of actions, like requesting, informing or acknowledging with a series of semantic notions known as slots. In the full dialog scenario, we refer these intents for an utterance within a dialog turn as dialog acts~\cite{abbeduto_1983}.
In Table \ref{tab:dialog_ex} as an example in Microsoft dialogue challenge dataset~\cite{li2018microsoft}, each utterance or response may involve more than one dialog acts with or without specific targeted slots.

% method: single utterance, single intent
% problem: multiple intents, intent fluidity
% solution: context
Within traditional SLU frameworks, such identification process is usually articulated as a single intent classification task coupled with a slot labeling task by dissecting a dialog into single utterances~\cite{liu2017, xia-etal-2018-zero}.
While most works induce large success in modeling the separate or joint distribution from intents and slots~\cite{goo-etal-2018-slot, li-etal-2018-self, e-etal-2019-novel, liu2019cmnet}, systems trained with such independent locutionary sentences quickly suffer from the insufficiency of capturing comprehensive semantics as the dialog flows. Irregularity and high mutability of user utterances may make it more difficult to capture precise user intents, especially for colloquial or implicit utterances~\cite{purohitshort}.
% 1. multi-intent
Moreover, in real world scenario, an utterance can be associated with more than one intent~\cite{goo-etal-2018-slot, qin2019stackpropagation, qin2020agif, Rash:19}. Dominant SLU systems have adopted several techniques to model such sophisticated semantic natures.~\cite{Rash:19} first explored the joint multi-intent and slot-filling task by treating multiple intents as a single context vector, but not scalable to a larger number of intents.~\cite{qin2020agif} proposed a SOTA model to exploit slot-intent relations with the graph attention.

% context related work
\begin{table}[t]
  \footnotesize
  \caption{Snippet of a single turn within dialog data conversation with corresponding dialog acts and slots.}
  \label{tab:dialog_ex}
  \centering
  \begin{tabular}{l|l}
    \toprule
    \textbf{Speaker} & \textbf{Utterance} \\
    \midrule
    \hline
    \textbf{1. system} & \tabincell{l}{I had 10 restaurants. 2g Japanese \\Brasserie is great for you.} \\
    \hline
    \textbf{Act}: \textit{Offer} & \textbf{Slots}: \textit{(name: 2g Japanese Brasserie)}\\
    \textbf{Act}: \textit{Inform\_count} & \textbf{Slots}: \textit{(count: 10)}\\
    \hline\hline
    \textbf{2. user} & \tabincell{l}{Yes, 2g Japanese works. \\ I want to reserve there.} \\
    \hline
    \textbf{Act}: \textit{Inform\_intent} & \textbf{Slots}: \textit{(reserve\_restaurant: True)} \\
    \textbf{Act}: \textit{Select} & \textbf{Slots}: \textit{(name: 2g Japanese)} \\
    \hline\hline
    \textbf{3. system} & \tabincell{l}{Please confirm the following details: \\ Booking a table at 2g Japanese \\ Brasserie. The city is San Francisco.} \\
    \hline
    \textbf{Act}: \textit{Confirm} & \tabincell{l}{\textbf{Slots}: \textit{(name:} \textit{2g Japanese Brasserie)} \\\textit{ (city: San Francisco)}} \\
    \hline\hline
    \textbf{4. user} & \tabincell{l}{Yes the restaurant schedule works \\  for me. Do they have live music?  \\ How pricey is it?} \\
    \hline
    \textbf{Act}: \textit{Affirm} & \textbf{Slots}: \textit{()} \\
    \textbf{Act}:  \textit{Request} & \tabincell{l}{\textbf{Slots}: \textit{(has\_live\_music: None)}, \\ \textit{(price\_range: None)}} \\
    % \hline\hline
    % \textbf{5. system} & \tabincell{l}{Your reservation was successful. \\ They do not have live music. \\ Their prices are moderate.} \\
    % \hline
    % \textbf{Act}: \textit{Inform} & \tabincell{l}{\textbf{Slots}: \textit{(has\_live\_music: False)}, \\ \textit{(price\_range: Moderate)}}\\
    % \textbf{Act}: \textit{Notify\_success} & \textbf{Slots}: \textit{()} \\
    % \hline\hline
    % \textbf{6. user} & Thank you! \\
    % \hline
    % \textbf{Act}: \textit{Thank\_you} & \textbf{Slots}: \textit{()} \\
    % \hline
    \bottomrule
  \end{tabular}

\end{table}

\begin{figure*}[t]
  \centering
  \includegraphics[width=0.8\linewidth]{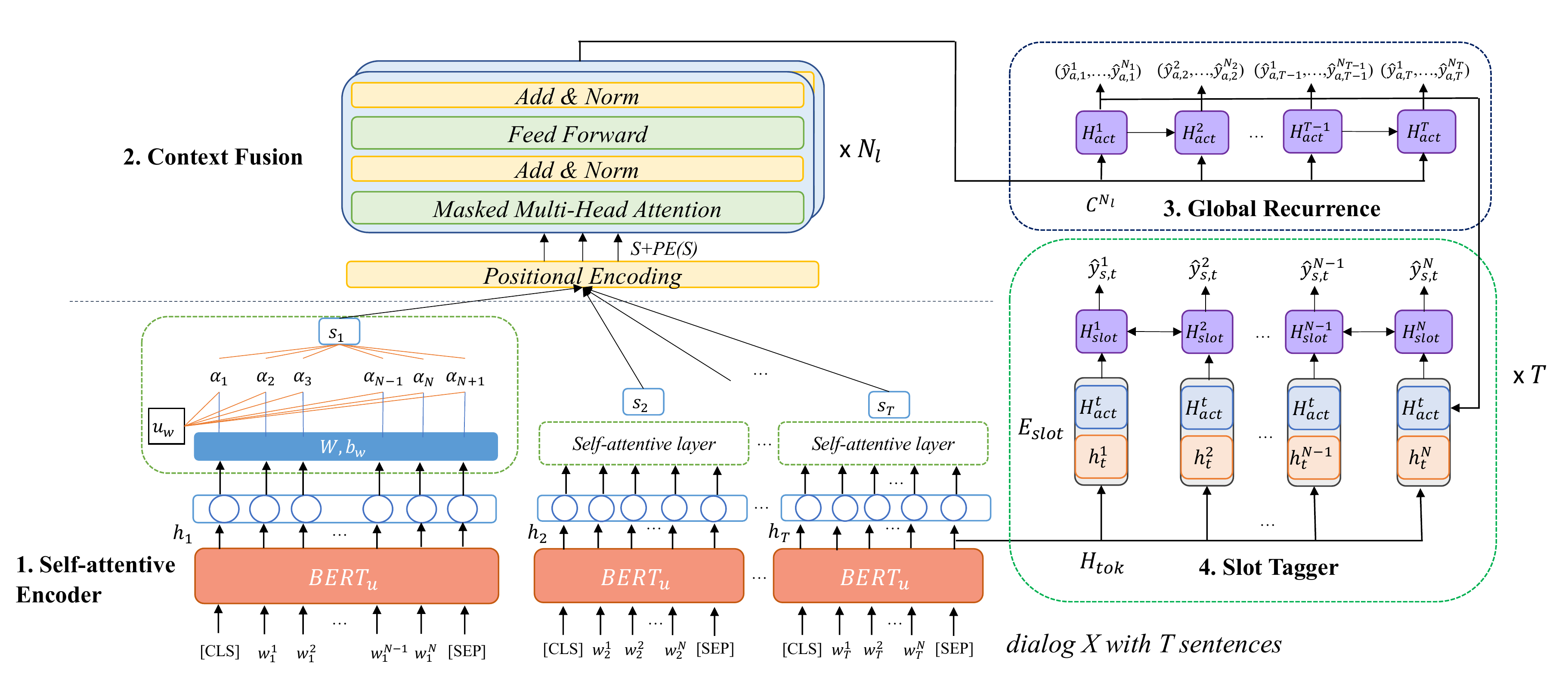}
  \caption{Illustration of our proposed framework for joint dialog act detection and slot filling in multi-turn dialogs.}
  \label{fig:model}
\end{figure*}

% 2. intent fluidity
% problem: error propagation
% why previous model does not consider: DST to model context
However, these approaches trained with independent utterances may not be sufficient in detecting contextual natures of dialog acts within dialogs, especially with the multiple intent cases~\cite{li2018microsoft, rastogi2019towards}. 
First, the sequential dependency between acts are obvious in most dialog cases regardless of domains. For instance, in Table \ref{tab:dialog_ex}, we can see `Select` usually comes after `Offer`, and `Select`'s slot is usually one of `Offer`'s slots.
Second, in less stylized conversations, they usually exhibit a broader open set of less bounded purposes, subject to arbitrary changes during turn transitions~\cite{xia2020composed}. For instance in utterance 4, user presents another requests for the price even after the confirmation is done, which may result from heuristics of the `Japanese Brasserie` restaurant name, which sounds expensive.
% For instance (example), multi dialog acts in a current utterance may be relevant to some aforementioned keywords within contexts. We can observe utterance 4 and 5 has both two acts while `Request` corresponds to `Inform` and `Affirm` corresponds to `Notify success`. 
Without such correlation matching, interpretability is undermined for joint tasks without any contextual information. Although in pipeline-based frameworks~\cite{zhang2020recent}, the resolution of contextual utterances is typically addressed in the next dialog management module (DM)~\cite{shan-etal-2020-contextual}, dissecting dialogs unnaturally may still open the door for significant cascade of errors by neglecting contexts.

To avert such error propagation, some end-to-end dialog systems \cite{wen-etal-2017-network, bordes2017learning} have been proposed to directly bypass the necessity of tracing dialog acts. Nevertheless, they lead to lower accuracies and tend to be opaque with less semantic transparency of dialog policies and states. Instead, \cite{bertomeu-etal-2006-contextual} studied contextual phenomena, thereby emphasizing the use of natural language contexts. \cite{shi2015contextual} introduced contextual signals to the joint intent-slot tasks. However, contextual information in their work was limited in terms of turn interactions. \cite{gupta2019casanlu} further proposed CASA-NLU to incorporate miscellaneous context signals until the current turn to jointly predict intents and slots. However, naive attention between historical utterances may lose track of sequential information traversing with the dialog progress. ~\cite{zhong2018globallocally} and \cite{shan-etal-2020-contextual} in dialog state tracking leveraged contexts to predict utterance-level slot values, which however may not be compatible with an unknown ontology such as restaurant names.

In this work, we present a context-aware hierarchical BERT fusion network (CaBERT-SLU) to exploit dialog history for joint tasks. 
% There are mainly four computational components in CaBERT-SLU, \textbf{BERT self-attentive encoder, context fusion encoder, global recurrent unit and slot tagger}. 
Simply, CaBERT-SLU will extract both utterance and turn-level information to identify multiple dialog acts and exploit a slot tagger to predict slots during the entire dialog. Our contributions are as follows and the code is available in https://github.com/waynewu6250/CaBERT-SLU.  
% It outperforms several single utterance baselines and the ablation study also demonstrates each module's effectiveness in our model. 

\noindent 1.  We propose CaBERT-SLU, which is the first attempt to consider previous dialog history for joint multiple dialog act and slot filling tasks, where previous SLU works usually isolate the utterances and only detect single dialog act only. \\
2. We demonstrate the effectiveness of context fusion attention in joint tasks with the ablation study and visualization. \\
3. Experimental results show that our model achieves SOTA performances over several competitive baselines.

%%%%%%%%%%%%%%%%%%%%%%%%%%%%%%%%%%%%%%%%%%%%%%%%%%%%%%%%%

\section{Methodology}

\begin{table*}[ht]
\caption{Main results for the joint task on two datasets. We report accuracy (ID Acc) for all intent exact match, F1 scores (ID F1) based on each intent calculation. We also report intent accuracy (IO Acc) and F1 score (IO F1) with models trained only on intent detection task. $\ddagger$ indicates that Stack-Prop can only predict single intent which we solely report its ID/IO Acc. $\dagger$ indicates the significant improvement of $p$-value $<$ 0.05 compared to the previous best contextual baseline \cite{gupta2019casanlu}.} % title of Table
\footnotesize
\centering % used for centering table
\hspace*{-0.7cm}
\begin{tabular}{l||c|c|c|c|c|c||c|c|c|c}
\toprule
\textbf{Dataset}
&\multicolumn{3}{|c}{\textbf{MDC}}
&\multicolumn{3}{|c}{\textbf{SGD}}
&\multicolumn{2}{||c}{\textbf{MDC}}
&\multicolumn{2}{|c}{\textbf{SGD}}\\
\midrule
\hline
Model & ID F1 & ID Acc & SL F1 & ID F1 & ID Acc & SL F1 & IO F1 & IO Acc & IO F1 & IO Acc \\
\hline
Stack-Prop$^\dagger$   ~\cite{qin2019stackpropagation}      & -     & 82.00 & 78.86 & -     & 83.51 & 89.24 & -     & 82.30 & -     & 83.75\\
Joint MID-SF ~\cite{Rash:19}                      & 86.18 & 75.75 & 70.92 & 86.33 & 85.11 & 78.21 & 85.48 & 75.41 & 90.71 & 84.97\\
AGIF         ~\cite{qin2020agif}                  & 89.63 & 80.18 & 78.87 & 91.96 & 85.41 & 86.68 & 90.73 & 74.12 & 92.41 & 76.11\\
ECA          ~\cite{chauhan20}                    & 87.88 & 77.84 & 69.20 & 93.65 & 93.11 & 80.00 & 87.75 & 77.66 & 95.78 & 92.45\\
\hline
BERT         ~\cite{devlin2019bert}               & 90.67 & 81.84 & 78.21 & 95.23 & 92.75 & 89.65 & 89.71 & 81.43 & 94.96 & 92.63\\
BERT+SA      ~\cite{lin2017structured}            & 89.91 & 80.63 & 78.19 & 95.32 & 92.99 & 90.01 & 89.73 & 81.64 & 94.95 & 92.73\\
% BERT+label-aware    & 89.95	& 81.55	& 79.00	& 95.03	& 92.98	& 90.11 \\
BERT+CASA-NLU   ~\cite{gupta2019casanlu}          & 90.98 & 82.17 & 78.16 & 97.07 & 95.24 & 90.46 & 90.25 & 80.91 & 96.72 & 94.84\\
\hline
CaBERT-SLU          & \textbf{91.26} & \textbf{83.05}$^\dagger$	& \textbf{79.64}$^\dagger$	& \textbf{99.14}$^\dagger$	& \textbf{98.59}$^\dagger$	& \textbf{95.71}$^\dagger$
& \textbf{90.81} & \textbf{82.69}$^\dagger$ & \textbf{98.92}$^\dagger$ & \textbf{98.24}$^\dagger$\\
\bottomrule

\end{tabular}
\label{table:main} % is used to refer this table in the text
\end{table*}

\subsection{Problem Statement}

Suppose we have a predefined dialog act label set $\mathcal{Y}^a$ and a slot set $\mathcal{Y}^s$, given a dialog $X = \{x_1, x_2, \dots, x_T\}$ of total $T$ user utterances and system responses, we would like to detect multiple dialog acts and slots for each $x_t$. For dialog act detection, we formulate it as a multi-label classification problem where for each $x_t$, we aim to find multiple dialog acts $(y_{a,t}^1, y_{a,t}^i,...,y_{a,t}^{N_t}), \forall y_{a,t}^i \in \mathcal{Y}^a$, and $N_t$ is the total number of dialog acts of the sample $x_t$. And for the slot filling task, for each $x_t = \{w_t^1, w_t^2, ..., w_t^N\}$ with total $N$ words, we wish to learn a parameterized mapping function to map input words into corresponding slot tags $(y_{s,t}^1, y_{s,t}^i,...,y_{s,t}^N), \forall y_{s,t}^i \in \mathcal{Y}^s$.

\subsection{BERT self-attentive encoder}

As shown in Figure \ref{fig:model}, our model consists of four functional units. We first encode each sentence $x_t = \{w_t^1, w_t^2, ..., w_t^N\}$ in a dialog $X$ with a BERT encoder $BERT_u$ to obtain token-level representations $\{h_t^1, h_t^2, ..., h_t^N\}$. BERT~\cite{devlin2019bert} is a multi-layer transformer-based encoder containing multi-head self-attention layers. It sufficiently extracts the contextualized information for each word token with respect to overall utterance. For a dialog with $T$ sentences, such $T$ token-level hidden representations will be sent into both the downstream context fusion encoder and the slot tagger to respectively predict dialog acts and slots. 

To further obtain the sentence representation $s_t$ of each utterance $x_t$ based on $\{h_t^1, h_t^2, ..., h_t^N\}$ and better consider the individual word importance, we follow the work in~\cite{lin2017structured} to use a self-attentive network. At each time step $i$ at sentence $x_t$, we first feed each token-level hidden state $h_t^i$ into an affine transformation ($W, b_w$), $\bar{h}^i_t = W h_t^i + b_w$. And we use equation \ref{eq:alpha} to obtain score $\alpha^i_t$.
\begin{align}
    \alpha^i_t &= \frac{e^{{\bar{h}{^i_t}}^T u_w}}{\sum_{j} e^{{\bar{h}{^j_t}}^T u_w}}
    \label{eq:alpha}
\end{align}
% where $\{h_t^1, h_t^i, ..., h_t^N\}$ are token-level representations of $h_t$, which are fed into an affine transformation ($W, b_w$) and output a set of keys $\{\bar{h}^1_t, ..., \bar{h}^N_t\}$. 
Then $\{\alpha^i_t\}$ represents the similarity scores between each $h_t^i$ and $K$ heads of learnable context vectors $u_w$ which indicate the global sentence views; for each head, we can get a sentence representation $s^h_t = \sum_i \alpha^i_t h^i_t$. Finally we will concatenate all the heads for the final representation $s_t$. 
% This will help identify the particular words' importance to overall sentence representation.

\subsection{Context fusion encoder}

After obtaining the final sentence representation $S = \{s_1, s_2, \dots, s_T\} \in \mathbb{R}^{T \times H_b}$ for a dialog, where $H_b$ is BERT hidden size, we combine $\{s_1, s_2, \dots, s_T\}$ with a unidirectional transformer encoder, which is devised to model the contextual relevance information throughout $T$ sentences in the dialog. This context fusion encoder contains a stack of $N_l$ layers. There are a masked multi-head self-attention sublayer (Attention) and a point-wise fully connected feed-forward network (FFN) as shown in equation \ref{eq:attention}. It will first project $S$ with weight matrices: $W^Q, W^K, W^V \in \mathbb{R}^{H_b \times H_a}$ to be $S^Q = S W^Q$, $S^K = S W^K$, $S^V = S W^V$. 
Then each of them will be separated into $h$ heads, with each head $i$ to be $H_i \in \mathbb{R}^{T \times (H_a/h)}$, $H_a$ is the hidden size for the attention module. These $H_i$ will be sent into the self-attention layer.

\begin{equation}
    Attention(H^Q_i,H^K_i,H^V_i) = softmax
    (\frac{H^Q_i(H^K_i)^T}{\sqrt{H_b}})H^V_i
\end{equation}

\begin{equation}
    FFN(x) = max(0,xW_1 + b_1)W_2 + b_2
    \label{eq:attention}
\end{equation}

Then for the entire context fusion layer, we can unfold each layer $l$ in equation \ref{eq:context}, where $PE(\cdot)$ denotes positional encoding function. Here we omit the residual layer and layer normalization layer in the equation, which exist in real implementation.

\begin{align}
    C^1 &= S + PE(S) \\
    \label{eq:context}
    C^l &= FFN(Attention(C^{l-1}, C^{l-1}, C^{l-1})
\end{align}

\subsection{Global Recurrent Unit}

We found out that the context fusion encoder could introduce the mutual interaction between each utterance, which may nevertheless be insufficient to capture the global sequential information as the dialog progresses. 
Thus, we apply an additional unidirectional LSTM layer upon the context fusion layer to supplement such global relations to obtain the final output states $H_{act} \in \mathbb{R} ^{T \times H_L}$, where $H_L$ is the hidden size of LSTM.

\begin{align}
    H_{act} = LSTM(C^{N_l})
\end{align}

Then we can generate the logits $\hat{y}_a = \sigma(H_{act} W_{act})$ by transforming $H_{act}$ with $W_{act} \in \mathbb{R}^{H_L \times |\mathcal{Y}^a|}$ and a sigmoid function $\sigma$. Finally we can have the dialog act detection objective as a binary cross entropy loss where $N_s$ is number of samples, $T$ is the max dialog length in samples and $|\mathcal{Y}^a|$ is the number of total dialog acts:
\begin{align}
    \mathcal{L}_a \coloneqq - &\sum_{i=1}^{N_s}\sum_{t=1}^{T}\sum_{a=1}^{|\mathcal{Y}^a|} (y^{(i,a)}_t log (\hat{y}^{(i,a)}_t) \nonumber \\
    &+ (1-y^{(i,a)}_t) log (1-(\hat{y}^{(i,a)}_t)) 
\end{align}

\subsection{Slot Tagger}

In addition to detecting dialog acts, we further detect the slots for each utterance in the dialog. Here we take the hidden representation $H_{tok} = \{h_t^1, h_t^2, ..., h_t^N\}$ from $BERT_u$ again. Then we concatenate $H_{tok}$ with the dialog act context information $H_{act}$ to be slot hidden states $E_{slot} = H_{tok} \oplus H_{act}$. Then we use another BiLSTM as the slot-filling tagger and generate the logits for each token. 
\begin{align}
    H_{slot} = BiLSTM(E_{slot}) \\
    \hat{y}_s = softmax(H_{slot} W_{slot})
\end{align}
Finally we can define the cross entropy loss as the objective:
\begin{align}
    \mathcal{L}_s \coloneqq - &\sum_{i=1}^{N_s}\sum_{t=1}^{T}\sum_{j=1}^{N}\sum_{s=1}^{|\mathcal{Y}^s|} (y^{(i,j,s)}_t log (\hat{y}^{(i,j,s)}_t))
\end{align}
The final joint objective will be formulated as $\mathcal{L}_\theta = \mathcal{L}_a + \mathcal{L}_s$.

%%%%%%%%%%%%%%%%%%%%%%%%%%%%%%%%%%%%%%%%%%%%%%%%%%%%%%%%%

\section{Experiments}

\subsection{Datasets}

\begin{table*}[ht]
\caption{Ablation study on different components of CaBERT-SLU. We report accuracy (ID Acc) for all intents exact match and F1 scores (ID F1) based on the individual intent calculation; SL F1 for slot-filling F1 scores. We also report intent accuracy (IO Acc) and F1 score (IO F1) with models trained solely on intent detection without slot filling. SA: self-attentive layer, CF: context fusion layer.} % title of Table
\label{tab:ablation}
\footnotesize
\centering % used for centering table
\hspace*{-0.7cm}
\begin{tabular}{l||c|c|c|c|c|c||c|c|c|c}
\toprule
\textbf{Dataset}
&\multicolumn{3}{|c}{\textbf{MDC}}
&\multicolumn{3}{|c}{\textbf{SGD}}
&\multicolumn{2}{||c}{\textbf{MDC}}
&\multicolumn{2}{|c}{\textbf{SGD}}\\
\midrule
\hline
Model & ID F1 & ID Acc & SL F1 & ID F1 & ID Acc & SL F1 & IO F1 & IO Acc & IO F1 & IO Acc\\
\hline
BERT                        & 90.67 & 81.84 & 78.21 & 95.23 & 92.75 & 89.65 & 89.71 & 81.43 & 94.96 & 92.63\\
\;+LSTM                     & 90.83 & 82.10 & 78.79 & 98.61 & 97.78 & 89.65 & 90.55 & 82.13 & 98.45 & 97.55\\
\;+SA+LSTM                  & 90.84 & 82.38 & 78.61 & 98.61 & 97.79 & 90.19 & 90.53 & 82.02 & 98.48 & 97.65\\
\;+SA+CF                    & 90.89 & 82.59 & \textbf{79.81} & 98.94 & 98.19 & 95.53 & \textbf{90.82} & 81.83 & 98.82 & 98.01\\
\hline
CaBERT-SLU  & \textbf{91.26} & \textbf{83.05}	& 79.64	& \textbf{99.14}	& \textbf{98.59}	& \textbf{95.71} 
                              & 90.81 & \textbf{82.69} & \textbf{98.92} & \textbf{98.24}\\
\bottomrule

\end{tabular}
\label{table:ablation} % is used to refer this table in the text
\end{table*}

\begin{table}[t]
  \caption{CaBERT-SLU performance in different domains.}
  \label{tab:domain}
  \footnotesize
  \centering
  \begin{tabular}{ll||c|c|c}
    \toprule
    \textbf{Dataset}      & \textbf{Domain} & ID F1 & ID Acc & SL F1       \\
    \midrule
    \multirow{3}{*}{\textbf{MDC}}&restaurant & 86.41	& 75.28	& 74.68 \\
    &taxi & \textbf{94.30} & \textbf{88.47} & 80.97 \\
    &movie & 93.02 & 85.74 & \textbf{82.35} \\
    \midrule
    \multirow{3}{*}{\textbf{SGD}}&restaurant & 98.92 & 98.30 & 94.15 \\
    &single & \textbf{99.14} & 98.31 & 91.27 \\
    &multiple & 99.10 & \textbf{98.60} & \textbf{95.92} \\
    \bottomrule
  \end{tabular}
\end{table}

We evaluate our model on two multi-turn dialog datasets: Microsoft Dialogue Challenge dataset (MDC) \cite{li2018microsoft} and Schema-Guided Dialogue dataset (SGD) \cite{rastogi2019towards}. \textbf{MDC} is a human-annotated conversation dataset in three domains (movie, restaurant, taxi). Each of them contains 2890, 4103, 3094 dialogs with total 11 acts and 50 slots. For each utterance, it is attached one or more dialog acts and slots. \textbf{SGD} consists of over 20k annotated multi-domain, task-oriented conversations between a human and a virtual assistant. These conversations span 20 domains, ranging from banks and events to media, travel and weather. We mainly adopt the user and system acts in each utterance for the dialog act detection and corresponding slots for slot filling. It has total 18 acts and 89 slots. We mainly split the training/validation/testing data with the ratio 0.7/0.1/0.2. 

% \begin{table}[t]
% \small
% \centering % used for centering table
% \begin{tabular}{|l|l|l|l|}
% \hline
% Dataset & train/val/test & \# Intents & \# Slots\\
% \hline
% MDC      & 45k/15k/15k & 11 & \\
% SGD      & 198k/66k/66k & 18 & \\
% \hline
% \end{tabular}
% \caption{Dataset statistics} % title of Table
% \label{table:data} % is used to refer this table in the text
% \end{table}

\subsection{Experimental Setup}

% \subsubsection{Baseline Models}

We compare our results with several competitive baselines:
\textbf{Stack-Prop}~\cite{qin2019stackpropagation} which uses two stacked encode-decoder structures for joint single intent and slot filling tasks. \textbf{Joint MID-SF}~\cite{Rash:19} which first considers multi-intent detection task in use of BiLSTMs, \textbf{AGIF}~\cite{qin2020agif} which uses graph interactive framework to consider fine-grained information, \textbf{ECA}~\cite{chauhan20} which encodes context with LSTM encoder for joint task prediction. We also fine-tune BERT pretrained layers with several proposed work including a self-attentive layer (\textbf{SA}) from~\cite{lin2017structured} and \textbf{CASA-NLU} which encodes context with DiSAN sentence2token~\cite{gupta2019casanlu}.
% label-aware layer which considers the label embedding interaction with single utterance embedding

% \subsubsection{Settings}

For experimental setting, we exploit the pretrained BERT model with 12 layers of 768 hidden units and 12 self-attention heads. For the self-attentive layer, we use 4 heads of context vectors. For the context fusion encoder, we set 6 transformer layers and the max sequence length as 60. Both two LSTMs have 256 hidden units. We use the batch size of 4 dialogs for MDC and 2 for SGD. In all training, we use Adam optimizer with learning rate as 2e-5. The model is trained for 20 epochs with the best performance on validation set. For metrics, by following \cite{qin2020agif}, we evaluate the performance of dialog act detection with accuracy and macro F1 score. We use F1 score for slot filling. Here we only consider a true positive when all BIO values for a slot is correct and forfeit `O` tags.

\section{Results}

\subsection{Main Results}

\begin{figure}[t]
  \centering
  \includegraphics[width=1.1\linewidth]{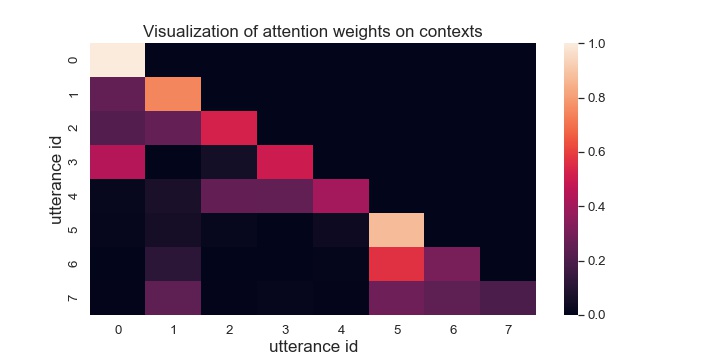}
  \caption{Visualization of attention weights on the last layer of context fusion encoder.}
  \label{fig:vis}
\end{figure}

Table \ref{table:main} shows the performance of CaBERT-SLU on joint tasks in two dialogue datasets, compared with several baseline models. Our model beats all baselines whether they are based on single utterances or BERT-related techniques, and achieves 1.1\% and 3.5\% increase in intent accuracy of two datasets than BERT+CASA-NLU~\cite{gupta2019casanlu}. We believe that the strong performance yielded by CaBERT-SLU pertains to the robust contextual information sharing both mutually and sequentially in multiple layers of masked self-attention. Without learning different weights of the dialog history to the current turn, previous approaches' performances are significantly undermined.  Also, it achieves 1.9\% and 5.8\% increase in slot F1 score, benefited from our model's contextual sharing. We can also observe a slight increase overall by considering the joint task.
We estimate the effectiveness of each module of CaBERT-SLU by conducting ablation experiments as shown in Table \ref{tab:ablation}. We observe a slight drop of 0.60\% without using self-attentive layer. And by incorporating both context fusion layer and global recurrence layer, it can boost the performance by overall roughly 1.5\% and almost 7\% in SGD slot filling task.

% \subsection{Multi-Domain evaluation}

To explore performances within different domains of our model, we separate MDC based on three domains. As for SGD, since a single dialog may involve multiple domains, we instead sample SGD with three variations: (1) dialogs associated with the `restaurant` domain (2) dialogs with only `single` domain (3) dialogs with `multiple` domains. In Table \ref{tab:domain}, we can observe that taxi and movie domains are much easier than restaurant domain which contribute more in joint scores. In SGD, we can see our method performs well regardless of how data is subsampled; especially outperforming on multi-domain dialogs. Our model also performs well in multiple domain for slot filling where context fusion may benefit domain transition of slots.

\subsection{Attention Visualization}

To further understand the mechanism of our context fusion encoder, we visualize the attention weights over the mean of heads at the last layer, as shown in Figure \ref{fig:vis}. In the example dialog, user first asks system to find a kid friendly place to eat. And the system asks about time, date and number of people at id 1 and 3.
Then we can observe id 2 and 4 have more weights on their close previous neighbors, which indicate the sequential relation between request and inform. We can see id 3 of asking number of people may be related to kid friendly keyword at id 0. After system talks about options of restaurants at id 5, id 6 replies with more dependency based on it. To note, with masked self-attention, we only attend weights on previous contexts.

% \section{Related Work}

\section{Conclusion}

In this work, we introduce an effective model that composes a contextual hierarchical structure affiliated with BERT to reinforce the connection between dialog contexts, which is often ignored by recent SLU works. By exploiting such naturalness of dialog flow, it is capable of capturing necessary mentions in previous dialog history for current tasks. Experimental results show that our model achieves strong improvements over models without contextual awareness. We also achieve SOTA results in joint multi-intent detection and slot filling of two multi-turn dialog datasets, without sacrificing the mutual relations between SLU and DM, and further error propagation.

% \section{Acknowledgements}

% The ISCA Board would like to thank the organizing committees of the past INTERSPEECH conferences for their help and for kindly providing the template files. \\
% Note to authors: Authors should not use logos in the acknowledgement section; rather authors should acknowledge corporations by naming them only.

\bibliographystyle{IEEEtran}

\bibliography{mybib}

\begin{thebibliography}{10}
\providecommand{\url}[1]{#1}
\csname url@samestyle\endcsname
\providecommand{\newblock}{\relax}
\providecommand{\bibinfo}[2]{#2}
\providecommand{\BIBentrySTDinterwordspacing}{\spaceskip=0pt\relax}
\providecommand{\BIBentryALTinterwordstretchfactor}{4}
\providecommand{\BIBentryALTinterwordspacing}{\spaceskip=\fontdimen2\font plus
\BIBentryALTinterwordstretchfactor\fontdimen3\font minus
  \fontdimen4\font\relax}
\providecommand{\BIBforeignlanguage}[2]{{%
\expandafter\ifx\csname l@#1\endcsname\relax
\typeout{** WARNING: IEEEtran.bst: No hyphenation pattern has been}%
\typeout{** loaded for the language `#1'. Using the pattern for}%
\typeout{** the default language instead.}%
\else
\language=\csname l@#1\endcsname
\fi
#2}}
\providecommand{\BIBdecl}{\relax}
\BIBdecl

\bibitem{zhang2020recent}
Z.~Zhang, R.~Takanobu, Q.~Zhu, M.~Huang, and X.~Zhu, ``Recent advances and
  challenges in task-oriented dialog system,'' 2020.

\bibitem{weld2021survey}
H.~Weld, X.~Huang, S.~Long, J.~Poon, and S.~C. Han, ``A survey of joint intent
  detection and slot-filling models in natural language understanding,'' 2021.

\bibitem{abbeduto_1983}
L.~Abbeduto, ``Linguistic communication and speech acts. kent bach robert
  m. harnish. cambridge: M.i.t. press, 1979, pp. xvii 327.'' \emph{Applied
  Psycholinguistics}, vol.~4, no.~4, p. 397–407, 1983.

\bibitem{li2018microsoft}
X.~Li, S.~Panda, J.~Liu, and J.~Gao, ``Microsoft dialogue challenge: Building
  end-to-end task-completion dialogue systems,'' \emph{arXiv preprint
  arXiv:1807.11125}, 2018.

\bibitem{liu2017}
T.~Liu, X.~DING, Y.~QIAN, and Y.~CHEN, ``Identification method of user's travel
  consumption intention in chatting robot,'' \emph{SCIENTIA SINICA
  Informationis}, vol.~47, p. 997, 08 2017.

\bibitem{xia-etal-2018-zero}
\BIBentryALTinterwordspacing
C.~Xia, C.~Zhang, X.~Yan, Y.~Chang, and P.~Yu, ``Zero-shot user intent
  detection via capsule neural networks,'' in \emph{Proceedings of the 2018
  Conference on Empirical Methods in Natural Language Processing}.\hskip 1em
  plus 0.5em minus 0.4em\relax Brussels, Belgium: Association for Computational
  Linguistics, Oct.-Nov. 2018, pp. 3090--3099. [Online]. Available:
  \url{https://www.aclweb.org/anthology/D18-1348}
\BIBentrySTDinterwordspacing

\bibitem{goo-etal-2018-slot}
\BIBentryALTinterwordspacing
C.-W. Goo, G.~Gao, Y.-K. Hsu, C.-L. Huo, T.-C. Chen, K.-W. Hsu, and Y.-N. Chen,
  ``Slot-gated modeling for joint slot filling and intent prediction,'' in
  \emph{Proceedings of the 2018 Conference of the North {A}merican Chapter of
  the Association for Computational Linguistics: Human Language Technologies,
  Volume 2 (Short Papers)}.\hskip 1em plus 0.5em minus 0.4em\relax New Orleans,
  Louisiana: Association for Computational Linguistics, Jun. 2018, pp.
  753--757. [Online]. Available:
  \url{https://www.aclweb.org/anthology/N18-2118}
\BIBentrySTDinterwordspacing

\bibitem{li-etal-2018-self}
\BIBentryALTinterwordspacing
C.~Li, L.~Li, and J.~Qi, ``A self-attentive model with gate mechanism for
  spoken language understanding,'' in \emph{Proceedings of the 2018 Conference
  on Empirical Methods in Natural Language Processing}.\hskip 1em plus 0.5em
  minus 0.4em\relax Brussels, Belgium: Association for Computational
  Linguistics, Oct.-Nov. 2018, pp. 3824--3833. [Online]. Available:
  \url{https://www.aclweb.org/anthology/D18-1417}
\BIBentrySTDinterwordspacing

\bibitem{e-etal-2019-novel}
\BIBentryALTinterwordspacing
H.~E, P.~Niu, Z.~Chen, and M.~Song, ``A novel bi-directional interrelated model
  for joint intent detection and slot filling,'' in \emph{Proceedings of the
  57th Annual Meeting of the Association for Computational Linguistics}.\hskip
  1em plus 0.5em minus 0.4em\relax Florence, Italy: Association for
  Computational Linguistics, Jul. 2019, pp. 5467--5471. [Online]. Available:
  \url{https://www.aclweb.org/anthology/P19-1544}
\BIBentrySTDinterwordspacing

\bibitem{liu2019cmnet}
Y.~Liu, F.~Meng, J.~Zhang, J.~Zhou, Y.~Chen, and J.~Xu, ``Cm-net: A novel
  collaborative memory network for spoken language understanding,'' 2019.

\bibitem{purohitshort}
H.~{Purohit}, G.~{Dong}, V.~{Shalin}, K.~{Thirunarayan}, and A.~{Sheth},
  ``Intent classification of short-text on social media,'' in \emph{2015 IEEE
  International Conference on Smart City/SocialCom/SustainCom (SmartCity)},
  2015, pp. 222--228.

\bibitem{qin2019stackpropagation}
L.~Qin, W.~Che, Y.~Li, H.~Wen, and T.~Liu, ``A stack-propagation framework with
  token-level intent detection for spoken language understanding,'' 2019.

\bibitem{qin2020agif}
L.~Qin, X.~Xu, W.~Che, and T.~Liu, ``Agif: An adaptive graph-interactive
  framework for joint multiple intent detection and slot filling,'' 2020.

\bibitem{Rash:19}
R.~Gangadharaiah and Balakrishnan, \emph{Joint multiple intent detection and
  slot labeling for goal-oriented dialog}.\hskip 1em plus 0.5em minus
  0.4em\relax Proc. of NAACL, 2019.

\bibitem{rastogi2019towards}
A.~Rastogi, X.~Zang, S.~Sunkara, R.~Gupta, and P.~Khaitan, ``Towards scalable
  multi-domain conversational agents: The schema-guided dialogue dataset,''
  \emph{arXiv preprint arXiv:1909.05855}, 2019.

\bibitem{xia2020composed}
C.~Xia, C.~Xiong, P.~Yu, and R.~Socher, ``Composed variational natural language
  generation for few-shot intents,'' 2020.

\bibitem{shan-etal-2020-contextual}
\BIBentryALTinterwordspacing
Y.~Shan, Z.~Li, J.~Zhang, F.~Meng, Y.~Feng, C.~Niu, and J.~Zhou, ``A contextual
  hierarchical attention network with adaptive objective for dialogue state
  tracking,'' in \emph{Proceedings of the 58th Annual Meeting of the
  Association for Computational Linguistics}.\hskip 1em plus 0.5em minus
  0.4em\relax Online: Association for Computational Linguistics, Jul. 2020, pp.
  6322--6333. [Online]. Available:
  \url{https://www.aclweb.org/anthology/2020.acl-main.563}
\BIBentrySTDinterwordspacing

\bibitem{wen-etal-2017-network}
\BIBentryALTinterwordspacing
T.-H. Wen, D.~Vandyke, N.~Mrk{\v{s}}i{\'c}, M.~Ga{\v{s}}i{\'c}, L.~M.
  Rojas-Barahona, P.-H. Su, S.~Ultes, and S.~Young, ``A network-based
  end-to-end trainable task-oriented dialogue system,'' in \emph{Proceedings of
  the 15th Conference of the {E}uropean Chapter of the Association for
  Computational Linguistics: Volume 1, Long Papers}.\hskip 1em plus 0.5em minus
  0.4em\relax Valencia, Spain: Association for Computational Linguistics, Apr.
  2017, pp. 438--449. [Online]. Available:
  \url{https://www.aclweb.org/anthology/E17-1042}
\BIBentrySTDinterwordspacing

\bibitem{bordes2017learning}
A.~Bordes, Y.-L. Boureau, and J.~Weston, ``Learning end-to-end goal-oriented
  dialog,'' 2017.

\bibitem{bertomeu-etal-2006-contextual}
\BIBentryALTinterwordspacing
N.~Bertomeu, H.~Uszkoreit, A.~Frank, H.-U. Krieger, and B.~J{\"o}rg,
  ``Contextual phenomena and thematic relations in database {QA} dialogues:
  results from a {W}izard-of-{O}z experiment,'' in \emph{Proceedings of the
  Interactive Question Answering Workshop at {HLT}-{NAACL} 2006}.\hskip 1em
  plus 0.5em minus 0.4em\relax New York, NY, USA: Association for Computational
  Linguistics, Jun. 2006, pp. 1--8. [Online]. Available:
  \url{https://www.aclweb.org/anthology/W06-3001}
\BIBentrySTDinterwordspacing

\bibitem{shi2015contextual}
Y.~Shi, K.~Yao, H.~Chen, Y.-C. Pan, M.-Y. Hwang, and B.~Peng, ``Contextual
  spoken language understanding using recurrent neural networks,'' April 2015.

\bibitem{gupta2019casanlu}
A.~Gupta, P.~Zhang, G.~Lalwani, and M.~Diab, ``Casa-nlu: Context-aware
  self-attentive natural language understanding for task-oriented chatbots,''
  2019.

\bibitem{zhong2018globallocally}
V.~Zhong, C.~Xiong, and R.~Socher, ``Global-locally self-attentive dialogue
  state tracker,'' 2018.

\bibitem{chauhan20}
S.~A. A. J. S.~S. Chauhan~A., Malhotra~A., ``Encoding context in task-oriented
  dialogue systems using intent, dialogue acts, and slots.'' in \emph{Saini H.,
  Sayal R., Buyya R., Aliseri G. (eds) Innovations in Computer Science and
  Engineering. Lecture Notes in Networks and Systems, vol 103. Springer,
  Singapore.}

\bibitem{devlin2019bert}
J.~Devlin, M.-W. Chang, K.~Lee, and K.~Toutanova, ``Bert: Pre-training of deep
  bidirectional transformers for language understanding,'' 2019.

\bibitem{lin2017structured}
Z.~Lin, M.~Feng, C.~N. dos Santos, M.~Yu, B.~Xiang, B.~Zhou, and Y.~Bengio, ``A
  structured self-attentive sentence embedding,'' 2017.

\end{thebibliography}

% \section{Appendix}
% \label{appendix:example}
%       \begin{table}[h!]
%       \caption{Dialog example of visualization in Figure \ref{fig:vis}.}
%       \label{tab:example}
%       \centering
%       \begin{tabular}{l|l|l}
%         \toprule
%         \textbf{Id}  & \textbf{Utterance} & \textbf{Act} \\
%         \midrule
%         0 & 
%         \tabincell{l}{Can you find me a good \\ american style kid friendly \\ place in Los Angeles, CA?} & 'request'
%         \\
%         \hline
%         1 & 
%         \tabincell{l}{I'm happy to assist you today! \\ 
%         What time and date would you \\ like your reservation for?}
%         & 'greeting'
%         \\
%         \hline
%         2 & 
%         \tabincell{l}{This is for Sunday around noon.}
%         & 'inform'
%         \\
%         \hline
%         3 & 
%         How many guests are in your party?
%         & \tabincell{l}{'request' \\'choice'}
%         \\
%         \hline
%         4 & 
%         2
%         & \tabincell{l}{'inform'}
%         \\
%         \hline
%         5 & 
%         \tabincell{l}{Please let me know which \\ one you would like to book:\\
%         Tamo's Hanter or Hot's kitchen?}
%         & \tabincell{l}{'request' \\ 'inform'}
%         \\
%         \hline
%         6 & 
%         Hot's kitchen would work.
%         & 'inform'
%         \\
%         \hline
%         7 & 
%         \tabincell{l}{Thank you for booking with us! \\Your reservation has been confirmed.} 
%         & \tabincell{l}{'thanks' \\ 'confirm'}
%         \\
        
%         \bottomrule
%       \end{tabular}
%     \end{table}

\end{document}